\documentclass[10pt, conference]{IEEEtran_IGSmod}
\usepackage{cite}
\usepackage{amsmath,amssymb,amsfonts}
\usepackage{algorithmic}
\usepackage{graphicx}
\usepackage{textcomp}
\usepackage{xcolor}
\usepackage{wrapfig}
\def\BibTeX{{\rm B\kern-.05em{\sc i\kern-.025em b}\kern-.08em
    T\kern-.1667em\lower.7ex\hbox{E}\kern-.125emX}}
\begin{document}

\title{Write on Paper and Get the Online Digital Trace:\newline A New Era for Handwriting
\thanks{German Federal Ministry of Education and Research (BMBF), French National Research Agency (ANR)}
}
\onecolumn
\author{\IEEEauthorblockN{
Florent Imbert\textsuperscript{1,5}, Yann Soullard\textsuperscript{2}, Eric Anquetil\textsuperscript{1}, 
Tanja Harbaum\textsuperscript{3}, Alexey Serdyuk\textsuperscript{3}, Fabian Kreß\textsuperscript{3}, 
\\ Tim Hamann\textsuperscript{4}, Peter Kämpf\textsuperscript{4}
}  
\IEEEauthorblockA{
\textsuperscript{1}IRISA, Université de Rennes, INSA Rennes, France \\  
\textsuperscript{2}IRISA, Université Rennes 2, France \\  
\textsuperscript{3}Karlsruhe Institute of Technology (KIT), Karlsruhe, Germany \\  
\textsuperscript{4}STABILO International GmbH, Heroldsberg, Germany \\  
\textsuperscript{5}LTU, Machine Learning Group, Luleå University of Technology, Sweden\\
\textit{yann.soullard@irisa.fr}
}  
}

\maketitle
\begin{abstract}
Capturing the digital trace of handwriting usually requires a specific stylus and a compatible substrate, be it a capacitive touchscreen, an ElectroMagnetic Resonance (EMR) tablet as used in Wacom systems or special paper. While writing on regular paper offers rich haptics, no latency and is well known for improving information retention, no low-cost and widely accepted, effective solution exists to digitize such a pen trace. The challenge is to accurately track the pen's trajectory without an external reference system while allowing unrestricted freedom of pen movement across a surface. We propose an innovative solution that combines a digital pen, advanced artificial intelligence algorithms, and adaptive AI techniques to reconstruct the digital trace of handwriting. Our approach integrates hardware development, focusing on a sensor-equipped pen, with software innovations to optimize trajectory reconstruction and processing in real time using an embedded AI. This work aims to advance the state-of-the-art in automated trace reconstruction of handwriting, enabling a seamless connection between traditional handwriting on paper and capturing the trace digitally.
\end{abstract}


\section{Introduction}
Handwriting on paper and on digital devices largely coexists without significant overlap, requiring a choice between writing on paper or using a digital pen with a tablet. Each one has its advantages. On the one hand, digital handwriting is widely used in many applications, such as note-taking or annotating documents allowing to share and store what is written. In learning handwriting, it allows a wide range of applications and exercises with appropriate feedback on the written trace produced by the student. On the other hand, writing on paper offers simplicity and availability. In addition, various research studies have demonstrated the significant benefits of handwriting in developing learning and knowledge acquisition for children, as compared to typing on a keyboard \cite{Mueller14,Oviatt12}.  

With advancements in electronic pens and intelligent software, these developments in equipment and systems offer the opportunity to bridge the gap between the two worlds and to benefit from what each one offers. In this paper, we present a complete hardware-software system combining an electronic pen with compatible firmware and drivers that can work with any support such as screen, tablet and paper. In this work, we focus on dealing with handwriting on paper which is the main challenge and the major novelty. Our proposal is based on: 1) a digital pen, called Digipen, developed by STABILO International GmbH and which is equipped with inertial sensors to track pen movements; 2) various strategies to deal with handwriting signals and the use of a deep learning AI system to reconstruct handwriting traces from the inertial sensors; 3) AI algorithm optimization techniques to compress the network and embed the technology into the pen. By distributing system complexity between hardware and software, our proposal aims to enable efficient and fast AI execution.  

\section{Digipen: A digital pen for trajectory tracking}

A growing fraction of electronic tablets is sold together with a matching electronic pen. However, a multitude of mutually incompatible technologies exists. Especially for low-cost tablets, which are attractive for schools, matching electronic pens are often not available. In addition, a clear majority of parents and teachers prefer that children learn handwriting on paper and not on a tablet. Therefore, a need exists for an electronic pen which writes on paper and can interact with all tablets.

\begin{wrapfigure}{l}{0.40\textwidth}
\centering \includegraphics[height=0.8cm]{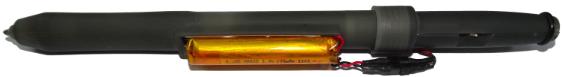}
\caption{Experimental pen with piggybacked battery.}
\label{fig:Digipen}
\end{wrapfigure}

The STABILO Digipen (Fig. \ref{fig:Digipen}) is a sensor-enhanced writing instrument with internal data processing capabilities and an external BLE (Bluetooth Low Energy) datalink for communication with mobile devices. It registers accelerations and rotation rates in three axes and the force on the writing tip. At the same time, it can be used as a regular pen, on regular paper. Therefore, it makes no special demands on the tablet outside of BLE connectivity and would be the ideal complement for any school tablets. 

Development was started with the goal of measuring handwriting proficiency in children. Next, we add handwriting recognition using software adapted from the well-known Kaldi toolkit, originally designed for speech recognition \cite{povey2011kaldi}. This matches the movement patterns with prerecorded patterns, which works well for experienced writers. However, since the pattern matching relies on overlearned movements, and children who are still in the process of learning handwriting cannot yet produce writing movement patterns with the required reliability, a different method of processing the data acquired by the pen is needed. Therefore, the development to reconstruct the pen trace only from inertial data by means of machine learning was started. Note that \cite{lunardini2020smart} previously proposed a similar pen based on IMU sensors but without the need to reconstruct the writing trace, which requires additional developments to acquire training samples.

\begin{figure}[h!]
\centerline{\includegraphics[height=3.6cm]{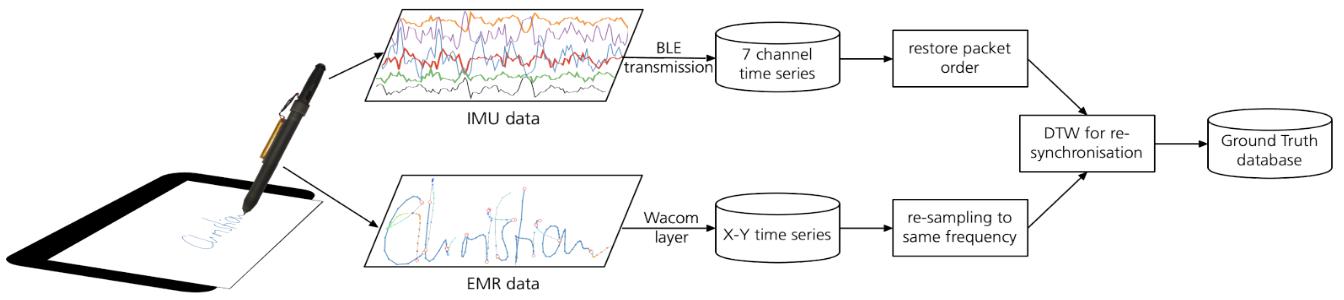}}
\caption{Processing pipeline to construct a ground truth database. Following \cite{Swaileh23}, a Dynamic Time Warping (DTW) alignment is applied to match IMU signal time frames with time frames from the tablet trajectory (EMR data).}
\label{fig:GTprocessing}
\end{figure}

To acquire the needed training data, we develop a special version of the Digipen to record both pen trace and inertial data at the same time. The electronics of Wacom inking pens is combined with an experimental version of the Digipen PCB, to be used on a tablet with Wacom’s recording layer and software. With a sheet of paper on the tablet, writing on paper and recording both inertial data and the pen trace as ground truth at the same time can be accomplished. Still, different sampling rates as well as delays in the transmission and processing of the BLE data required re-synchronizing the recordings of the Wacom software and the inertial data. This is achieved by re-sampling and using the onset of the tip force signal of both recording types as a clapperboard (Fig. \ref{fig:GTprocessing}). 

\section{Deep neural networks for handwriting trace reconstruction from IMU sensors }

Handwriting trace reconstruction from online handwriting has become a key focus in many systems, including pen-and-tablet setups and camera-based configurations. While offline to online transition can be a solution \cite{Elmokhtar23}, it often results in the loss of complex ductus, the intricate dynamics of strokes that characterize handwriting. Many tools able to capture digital handwriting traces are based on the use of a specific pen on a tablet, and it does not allow writing on paper. Camera-based tools like the Anoto pen \cite{Liao08} digitize handwriting using specialized microdot-patterned paper, but this approach is limited by the need for specific physical paper and the cost of the stylus. In this context, Inertial Measurement Unit (IMU) sensors are gaining traction as an affordable, low-power and versatile alternative, despite challenges associated with their inherently noisy signal. IMUs have demonstrated compatibility with digital tablets and even traditional pen-and-paper handwriting setups.  

Here, we address the task of reconstructing handwriting traces from IMU signals by writing with the Digipen on paper. We explore various strategies to deal with handwriting on paper. Inspired by previous works \cite{Swaileh23, Imbert25}, we benefit from a complete pipeline including advanced preprocessing and a mixture-of-experts' approach based on deep learning models to reconstruct digital handwriting traces by writing on paper.   

\subsection{Related Works}

Only a few papers have explored handwriting trace reconstruction from IMU signals. Wehbi et al. \cite{Wehbi22} extend the monowriter approach proposed by Ott et al. \cite{Ott22} to deal with multiple writers. This approach, based on deep learning techniques, uses handwriting on a tablet from an earlier version of the STABILO Digipen (version 5.5). In this work, the authors introduce a Convolutional Neural Network architecture designed for handwriting trace reconstruction, with linear interpolation to align sensor data with tablet-based ground truth for network training. 

More recently, \cite{Swaileh23} proposed a complete pipeline for handwriting trace reconstruction from Digipen data.  This pipeline relies on a new pre-processing including a Dynamic Time Warping alignment to align sensor data with tablet-based ground truth for network training and a neural network model inspired by Temporal Convolutional Networks (TCNs). Based on this work, we previously proposed a mixture of experts' approach that considers the specific characteristics of the signals \cite{Imbert25}. In this work, two expert models are designed, one dedicated to touching strokes and one for pen-up movements. In another work, a domain adaptation technique based on a Domain-Adversarial Neural Networks has been explored to process children's handwriting \cite{Imbert24}, exhibiting varying levels of graphomotor gesture development, including differences in fluency and speed. Until now, all the experiments were done by writing on a tablet.  

\subsection{Dealing with handwriting on paper}

Input signals vary significantly depending on whether the handwriting is captured on paper or on a tablet screen, due to differences in surface properties and friction. Paper typically offers higher friction, providing more resistance to the motion of a pen and causing additional noise in the IMU signals compared to handwriting on a tablet. This strongly impacts the performance of a handwriting reconstruction system trained only on data from tablets.  

To reconstruct the handwriting trace, we focus on the Mixture-Of-Experts approach based on Temporal Convolutional Networks which has been successfully used with input signals on data acquired on tablets using the Digipen \cite{Imbert25}. As in \cite{Imbert25}, we will refer to this model as MOE-CI. Here, we propose to compare different strategies to study the impact of noise and the differences between handwriting signals from tablet or paper. First, to benefit from the collection of tablet data \cite{Swaileh23, Imbert25}, we apply a Kalman filter to the signals from the pen on tablet in an attempt to reduce the noise and then evaluate the reconstruction using an MOE-CI model trained on tablet data. The second strategy consists in training an MOE-CI model from scratch on data acquired on paper using the dual acquisition process discussed above. We will compare those two strategies to the reference model from \cite{Imbert25}, trained only on handwriting acquired on a tablet and evaluate the paper data on it.  

\section{Hardware-aware Neural Network Compression}

The designed MOE-CI network enables handwriting trajectory reconstruction from IMU-data, however, to perform real-time inference on resource-constrained devices such as electronic pens, the neural network must be compressed for the target hardware. The current MOE-CI model is composed of two expert models. In total it requires 3.6 MB of ROM, 288 KB of RAM and 90 MFLOP for a single inference, whereas a typical system-on-chip suitable for an electronic pen has maximum 1 MB of ROM and 256 KB RAM in total. 

\begin{figure}[htbp]
\centerline{\includegraphics[height=6.cm]{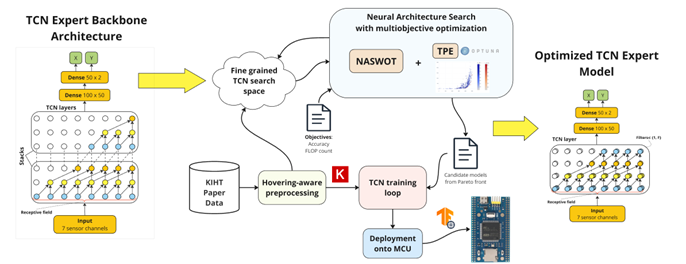}}
\caption{Proposed hardware-aware network optimisation concept.}
\label{fig:opt}
\end{figure}

To meet these hardware constraints, we conducted the search for the resource optimized yet still capable expert TCN models using our hardware-aware Neural Architecture Search (NAS) approach (Fig. \ref{fig:opt}). For that, we introduced expressive search space for the TCN-based models, which allows to vary the number of convolutional filter blocks of each individual layer besides the standard parameters of the TCNs such as kernel size, dilation factor, number of stacks and number of layers per stack. To efficiently sample architectures from the resulting search space, we apply the NASWOT zero-cost proxy \cite{Mellor21} instead of full NN training, since the NASWOT score correlates with the final accuracy of the trained network and at the same time can be evaluated quickly. The search of architectures was performed using the methods of multiobjective Bayesian optimization with NASWOT score maximization and FLOP count minimization goals. We trained several architectures from the resulting Pareto front and evaluated them in hardware. The best discovered architecture requires 10,952 parameters against 467,452 of the initial TCN expert model and as little as 74 KB RAM and 66 KB ROM, pertaining comparable trajectory reconstruction accuracy and allowing to deploy these networks onto the tiny microcontroller of the Digipen. 

\section{Results}

\begin{figure}[t!]
\centerline{\includegraphics[height=6.cm]{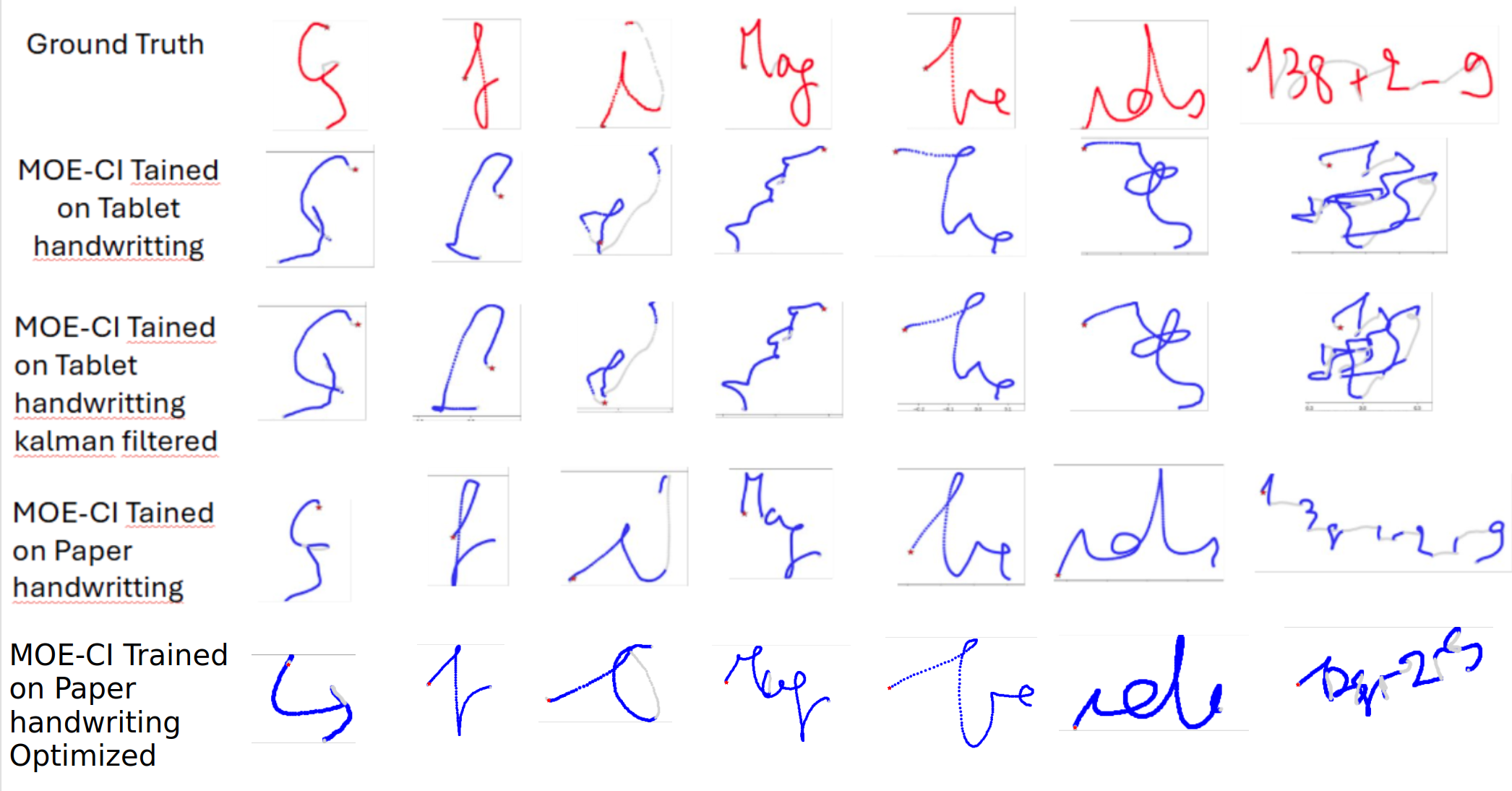}}
\caption{Examples of ground truth traces and reconstructed traces for different models and by applying the various strategies proposed in this work.}
\label{fig:results}
\end{figure}

Following \cite{Swaileh23}, we evaluate the models and the various strategies by computing the Fréchet distance. We present quantitative and qualitative results in  Table \ref{tab:results} and Fig. \ref{fig:results}. The smaller the mean Fréchet distance, the more accurately the reconstruction matches the ground truth.  Signal differences depending on the acquisition medium (tablet or paper) lead to a considerable degradation in accuracy, underscoring the model's difficulty in generalizing to differently noisy data. This is illustrated by applying an MOE-CI model trained only on tablet data compared to a specific training on data from paper. 

To address this, the first proposal is to apply a Kalman filter to the signals, aiming to reduce noise. Although the Kalman filter succeeded in smoothing the data and mitigating high-frequency noise, the improvement in mean Fréchet distance is marginal and insufficient to bridge the gap between tablet and paper data. A noise reduction process may not be sufficient to account for the inherent differences in writing media.

The second  strategy consisting in training a model on paper data significantly reduced the mean Fréchet distance, yielding notably better performance in reconstruction. This result underlines the importance of optimizing a model on data from the target acquisition medium, and adapting it to the specific characteristics and noise of the input data. 

\begin{table}[t!]
    \centering
    \caption{Mean Fréchet Distance ($\downarrow$) Between Reconstructed Trajectory and Ground Truth.}
    \label{tab:results}
    \begin{tabular}{|c|c|c|c|c|}
        \hline
        Model & \multicolumn{3}{c|}{MOE-CI} & Compressed MOE-CI \\
        \hline
        Acquisition medium & Tablet screen & Tablet screen & Paper & Paper \\
        \hline
        Filter & $\times$ & \textit{Kalman filter} & $\times$ & $\times$ \\
        \hline
        Fréchet Distance & 0.566 & 0.542 & \textbf{0.429} & 0.738 \\
        \hline
    \end{tabular}
\end{table} 

The hardware-aware neural network compression proposal results in a significant increase in the mean Fréchet distance, showing a degradation in trajectory reconstruction. However, the trace reconstruction when the pen touches the paper is quite close to that of the non-optimized model. It is primarily the next stroke repositioning in case of pen-up movement that suffers from the compression. This may be related to the expert model dedicated to the pen-up movement, which is trained on the entire sequence (including the touching parts) to enhance its reconstruction capabilities. In the compressed version, this could provide information that is not useful for pen-up movement reconstruction and affect performance. This highlights, however, that the compressed model remains effective in handling data written on paper.


\section*{Acknowledgment}

This work was funded by the German Federal Ministry of Education and Research (BMBF) under grant number 01IS21097A (KIHT).  This project is financed by the KIHT French-German bilateral ANR project and these four partners, IRISA (France), KIT (Germany), Learn\&Go (France) and STABILO (Germany). This work was performed using HPC resources from GENCI IDRIS (Grant 2021-AD011013148)


\begin{thebibliography}{00}
\bibitem{Mueller14} P.A. Mueller and D.M. Oppenheimer, ``The pen is mightier than the keyboard: Advantages of longhand over laptop note taking,'' Psychological science, 25(6), 1159-1168, 2014
\bibitem{Oviatt12} S. Oviatt, A. Cohen, A. Miller, K. Hodge, and A. Mann, ``The impact of interface affordances on human ideation, problem solving, and inferential reasoning,'' ACM Transactions on Computer-Human Interaction (TOCHI), vol. 19, 2012 
\bibitem{povey2011kaldi} D. Povey, A. Ghoshal, G. Boulianne, L. Burget, O. Glembek, N. Goel, M. Hannemann, P. Motlicek, Y. Qian, P. Schwarz, and others, ``The Kaldi speech recognition toolkit,'' in IEEE 2011 Workshop on Automatic Speech Recognition and Understanding, 2011.
\bibitem{lunardini2020smart} F. Lunardini, D. Di Febbo, M. Malavolti, M. Cid, M. Serra, L. Piccini, A. L. G. Pedrocchi, N. A. Borghese, and S. Ferrante, ``A smart ink pen for the ecological assessment of age-related changes in writing and tremor features,'' IEEE Transactions on Instrumentation and Measurement, vol. 70, pp. 1--13, 2020.
\bibitem{Elmokhtar23} M. M. Elmokhtar, T. Lelore and H. Mouchère, ``Point to segment distance DTW for online handwriting signals matching'', in 12th International Conference on Pattern Recognition Applications and Methods (ICPRAM), Portugal: SCITEPRESS - Science and Technology Publications, pp. 850–855, 2023. 
\bibitem{Liao08} C. Liao, F. Guimbretière, K. Hinckley, and J. Hollan, ``Papiercraft: A gesture-based command system for interactive paper'', in ACM Transactions on Computer-Human Interaction (TOCHI), 14(4), 1-27, 2008.
\bibitem{Wehbi22} M. Wehbi, D. Luge, T. Hamann, J. Barth, P. Kaempf, D. Zanca and B.M. Eskofier, ``Surface-free multi-stroke trajectory reconstruction and word recognition using an imu-enhanced digital pen'', Sensors, 22(14), 5347, 2022.
\bibitem{Ott22} F. Ott, D. Rugamer, L. Heublein, B. Bischl and C. Mutschler, ``Joint classification and trajectory regression of online handwriting using a multi-task learning approach,''  In Proceedings of the IEEE/CVF winter conference on applications of computer vision, pp. 266-276, 2022.
\bibitem{Mellor21} J. Mellor, J. Turner, A. Storkey, and E. J. Crowley, ``Neural Architecture Search without Training,'' In International conference on machine learning (PMLR), pp. 7588-7598, June 2021.
\bibitem{Imbert25} F. Imbert, E. Anquetil, Y. Soullard, R. Tavenard, ``Mixture-of-experts for handwriting trajectory reconstruction from IMU sensors'', Pattern Recognition, Volume 161, 2025.
\bibitem{Swaileh23} W. Swaileh, F. Imbert, Y. Soullard, R. Tavenard and E. Anquetil, ``Online handwriting trajectory reconstruction from kinematic sensors using temporal convolutional network,'' International Journal on Document Analysis and Recognition (IJDAR), 26(3), 289-302, 2023.
\bibitem{Imbert24} F. Imbert, R. Tavenard, Y. Soullard and  E. Anquetil, ``Domain Adaptation for Handwriting Trajectory Reconstruction from IMU Sensors,'' In International Conference on Document Analysis and Recognition, pp. 3-11, Automatically Domain-Adapted and Personalized Document Analysis (ADAPDA) workshop. Cham: Springer Nature Switzerland, 2024.
\end{thebibliography}
\end{document}